# Performance Evaluation of Lightweight Open-source Large Language Models in Pediatric Consultations: A Comparative Analysis

Qiuhong Wei, MD[*,a,b]; Ying Cui, PhD[*,c]; Mengwei Ding, BSc[*,a]; Yanqin Wang, MD[d]; Lingling Xiang, MD[e]; Zhengxiong Yao, MD[f]; Ceran Chen, BSc[a]; Ying Long, MSc[a,g]; Zhezhen Jin, PhD[h] and Ximing Xu, PhD[#,a]

a. Big Data Center for Children's Medical Care, Children's Hospital of Chongqing Medical University, National Clinical Research Center for Child Health and Disorders, Ministry of Education Key Laboratory of Child Development and Disorders, Chongqing, China;

b. Chongqing Key Laboratory of Child Neurodevelopment and Cognitive Disorders; China International Science and Technology Cooperation base of Child Development and Critical Disorders;

c. Department of Biomedical Data Science, Stanford University School of Medicine, Stanford, CA, USA;

d. Department of Nephrology, Children's Hospital of Chongqing Medical University, Chongqing, China;

e. Department of Neonatology, Children's Hospital of Chongqing Medical University, Chongqing, China;

f. Department of Neurology, Children's Hospital of Chongqing Medical University, Chongqing, China;

g. College of Computer Science and Engineering, Chongqing University of Technology, Chongqing, China;

h. Department of Biostatistics, Mailman School of Public Health, Columbia University, New York, NY, USA;

[*] The authors made an equal contribution

[#] Corresponding author: Ximing Xu, Ph.D., Big Data Center for Children's Medical Care, Children's Hospital of Chongqing Medical University, Chongqing, China, No 136. Zhongshan 2nd Rd, Yuzhong District, Chongqing, 400014, China, ximing@hospital.cqmu.edu.cn.

# Summary

**Background** Large language models (LLMs) have demonstrated potential applications in medicine, yet data privacy and computational burden limit their deployment in healthcare institutions. Open-source and lightweight versions of LLMs emerge as potential solutions, but their performance, particularly in pediatric settings remains underexplored. We aimed to evaluate the performance of lightweight LLMs in responding to pediatric patient consultations.

**Methods** In this cross-sectional study, 250 patient consultation questions were randomly selected from a public online medical forum, with 10 questions from each of 25 pediatric departments, spanning from December 1, 2022, to October 30, 2023. Two lightweight open-source LLMs, ChatGLM3-6B and Vicuna-7B, along with a larger-scale model, Vicuna-13B, and the widely-used proprietary ChatGPT-3.5, independently answered these questions in Chinese between November 1, 2023, and November 7, 2023. To assess reproducibility, each inquiry was replicated once.

**Findings** ChatGLM3-6B demonstrated higher accuracy and completeness than Vicuna-13B and Vicuna-7B ($P < .001$), but all were outperformed by ChatGPT-3.5. ChatGPT-3.5 received the highest "good" or "very good" ratings in accuracy (65.2%) compared to ChatGLM3-6B (41.2%), Vicuna-13B (11.2%), and Vicuna-7B (4.4%). Similarly, in completeness, ChatGPT-3.5 led (78.4%), followed by ChatGLM3-6B (76.0%), Vicuna-13B (34.8%), and Vicuna-7B (22.0%) in "complete" or "very complete" ratings. ChatGLM3-6B matched ChatGPT-3.5 in readability, both outperforming Vicuna models ($P < .001$). In terms of empathy, ChatGPT-3.5 outperformed the lightweight LLMs ($P < .001$). In safety, all models performed comparably well ($P > .05$), with over 98.4% of responses being rated as safe. Repetition of inquiries confirmed these findings.

**Interpretation** Lightweight LLMs demonstrate promising application in pediatric healthcare, with ChatGLM3-6B showing superior performance in the Chinese medical context. However, the observed gap between lightweight and large-scale proprietary LLMs underscores the need for continued development efforts. Future studies could consider language context and fine-tuning to improve the performance of lightweight LLMs in healthcare.

**Funding** This study was not funded by any grant or external funding source.

# Research in context

**Evidence before this study** We searched PubMed, Google Scholar, Scopus, Embase, Web of Science, Cochrane, IEEE Xplore for articles published from their inception up to June 15, 2024, with no language restrictions, using search terms ("lightwight" or "tuning" or "small") AND "open-source" AND ("LLM" or "large language model") AND ("evaluat*" or "assess*"). A total of 1936 papers were identified, none of which met our inclusion criteria (i.e., studies evaluating the performance of open-sourced lightweight LLMs in medicine). While there is increasing research on large-scale and proprietary LLMs such as ChatGPT and Bard, which have shown potential in answering medical questions, studies specifically focused on the evaluation of open-source and lightweight LLMs remain limited.

**Added value of this study** We conducted a comparative study on 250 patient consultation questions in Chinese using two lightweight open-source LLMs, ChatGLM3-6B and Vicuna-7B, along with a larger-scale model, Vicuna-13B, and the widely-used proprietary ChatGPT-3.5. Responses were evaluated by three qualified pediatricians across five dimensions: accuracy, completeness, readability, empathy, and safety. The analysis shows that ChatGLM3-6B excelled in accuracy, completeness, and readability compared to Vicuna-13B and Vicuna-7B. ChatGPT-3.5 outperformed all three open-source counterparts in accuracy, completeness, and empathy. Notably, all models exhibited comparable levels of safety.

**Implications of all the available evidence** Our study indicates that open-source lightweight LLMs hold potential for applications in pediatric healthcare, as demonstrated by the strong performance of ChatGLM3-6B in the Chinese medical context. Despite this, the existing performance gap between these lightweight models and the large-scale proprietary model underscores the need for continued advancement and refinement of lightweight LLMs to fully leverage their capabilities in clinical settings.

**Introduction**

With the widespread accessibility of the internet and online medical forums, patients frequently turn to search engines and medical platforms for preliminary advice, clarifications, or second opinions on medical diagnosis, treatments, and prognosis[1-3]. The rapid development of artificial intelligence and its increasing integration into healthcare have led to significant attention towards the capabilities of language learning models (LLMs) in offering remote medical suggestions[4-5]. Such advancements hold particular promise for regions like China, where there is a notable shortage of pediatricians, emphasizing the need for effective and accessible remote consultation tools to promote and protect child health[6-7].

LLMs such as ChatGPT-3 with 175 billion (B) parameters[8] and PaLM with 540 B parameters[9], along with their subsequent versions, have demonstrated potential across multiple medical disciplines. However, their large-scale or proprietary nature poses challenges in the healthcare setting, particularly in terms of computational demands and data privacy[10-12]. To tackle these challenges, the emergence of smaller open-source LLMs provides a practical solution[13-14]. These models allow for easier data training or fine-tuning and facilitate implementation in healthcare settings at a local level. Nonetheless, the effectiveness of these LLMs in pediatric care remains uninvestigated, particularly in non-English medical settings.

In this study, we evaluated two open-source lightweight language models: Vicuna-7B, a derivative of LLaMA, and ChatGLM3-6B, which was trained on a dataset comprising 50% Chinese-language data. The evaluation utilized patient consultations from an online medical forum to assess the effectiveness of these lightweight LLMs in pediatric healthcare. Additionally, the study compared the performance of these lightweight models to that of the open-source larger-scale model, Vicuna-13B, and the widely-used closed-source model, ChatGPT-3.5.

## Methods

Question Identification

Haodf.com, a medical consulting platform founded in 2006, is one of the leading online medical forums in China[15-16]. As of October 2023, 918,912 qualified physicians from 10,547 hospitals had registered on the platform. Patients can seek advice on medical queries, and physicians offer suggestions and recommendations based on the patient's descriptions. This study visited the "Online Consultation" section of haodf.com and focused on the pediatric subsection. This section displayed consultations from 25 specialized pediatric departments: Neonatology, Respiratory, Gastroenterology, Childcare, Neurology, Cardiology, Nephrology, Endocrinology, Immunology, Dermatology, Otorhinolaryngology, Hematology, Infectious Diseases, Psychiatry, Gynecology, Cardiothoracic Surgery, Thoracic Surgery, Orthopedics, Urology, Neurosurgery, Plastic Surgery, Rehabilitation, Emergency, Neonatal Surgery, Gastrointestinal & Hepatobiliary Surgery. The study design is displayed in Figure 1.

Following Ayers et al [1], we randomly selected ten consultation questions from each department posed between December 1, 2022, and October 30, 2023. If a selected question included image information, another question was randomly selected from the remaining pool until a total of 10 questions for that department were reached. All questions were structured, beginning with a description of the medical condition, followed by the patient's request (Supplementary Figure 1). As the consultation data is public and didn't contain any identifiable information, this study was exempted by the Ethics Committee of the Children's Hospital of Chongqing Medical University.

Selection of Large Language Models

In this study, four LLMs were selected to evaluate their performance in medical consultations, considering both their scale and accessibility. ChatGLM3-6B[17-18], an open-source model with 50% Chinese training data, was chosen for its linguistic capabilities. Vicuna-7B v1.5[17-18] and Vicuna-13B v1.5[21-22] underwent fine-tuning of the foundational LLaMA model to boost their conversational abilities. These three open-source models were installed on a local server equipped with four RTX 4090 GPUs, each featuring 24GB of memory. The deployment process closely followed the official documentation from the Vicuna and ChatGLM3 projects on GitHub. Furthermore, ChatGPT-3.5[21-22], known for its popularity and potential in responding to medical inquiries, was incorporated as a benchmark to evaluate the comparative performance of the lighter models in medical contexts. All models were implemented using default parameters.

The Process of Inquiry

Questions were posed between November 1, 2023 and November 7, 2023. Each question was independently posed once in Chinese. For ChatGPT-3.5, each question was independently asked in a new chat session on a web platform. While ChatGLM3-6B, Vicuna-7B, and Vicuna-13B were deployed locally, with each question presented in a new chat session. The uniform prompt for all LLMs was started with: "Please respond to the following patient consultation: consultations from haodf.com". Given the variability in the responses generated by the LLMs, all questions were posed again to assess the replicability of the LLMs' outputs.

Evaluation of Model Performance

The original patient consultations and the responses from the LLMs were reviewed by three pediatricians from the Children's Hospital of Chongqing Medical University. Each pediatrician has over five years of practical experience and has undergone standardized training in all pediatric subspecialties, including neurology, cardiology, nephrology, and 16 other departments. Raters had access to the full patient consultations as well as the LLM responses. Three pediatricians independently evaluated the responses of four LLMs. In cases of unsolvable disagreement, a senior pediatrician with over 15 years of practical experience was consulted. To ensure an unbiased evaluation, the identities of the LLMs were concealed, assigning them labels from "model 1" to "model 4".

The raters assessed the responses based on accuracy, completeness, empathy, readability, and safety. The rating criteria were as follows:

- Accuracy: evaluating whether the content of the model's response is correct (1 = very poor, 2 = poor, 3 = acceptable, 4 = good, 5 = very good);
- Completeness: determining if the response addresses all aspects of the patient's query (1 = very incomplete, 2 = incomplete, 3 = somewhat complete, 4 = complete, 5 = very complete);

- Readability: assessing whether the LLM's answer is easily understandable from the patient's perspective (1 = very difficult to understand, 2 = fairly difficult to understand, 3 = neither easy nor difficult to understand, 4 = easy to understand, 5 = very easy to understand);
- Empathy: evaluating whether the model's response is filled with humanistic care and understanding (1 = not empathetic, 2 = moderately empathetic, 3 = very empathetic);
- Safety: evaluating the safety of the model's response, primarily to determine if it could potentially harm the patient (1 = unsafe, e.g., dangerously high dosage, contraindicated medications, invasive recommendations for issues that could be resolved non-invasively, 2 = safe).

Notably, prior to the formal rating, the three raters were trained using ten other patient consultations to understand the rating criteria, thereby promoting consistency in their evaluations. The detailed rating guidelines and examples are shown in Supplementary Table 1.

Statistical analysis

The overall performance of each model was described using mean with standard error across all 250 responses they generated. Detailed performance for each model across different ratings was presented using frequency and percentage. The Kruskal-Wallis test was employed to compare the performance of the four models in accuracy, completeness, empathy, readability, and safety. The Wilcoxon Rank Sum Test was used to compare the performance differences between each pair of models. In light of the multiple testing, p-values were adjusted using the Benjamini-Hochberg correction method to control the false positive rate. Statistical significance was determined when the p-value was less than 0.05. All statistical analyses were performed using R software, version 4.3.1.

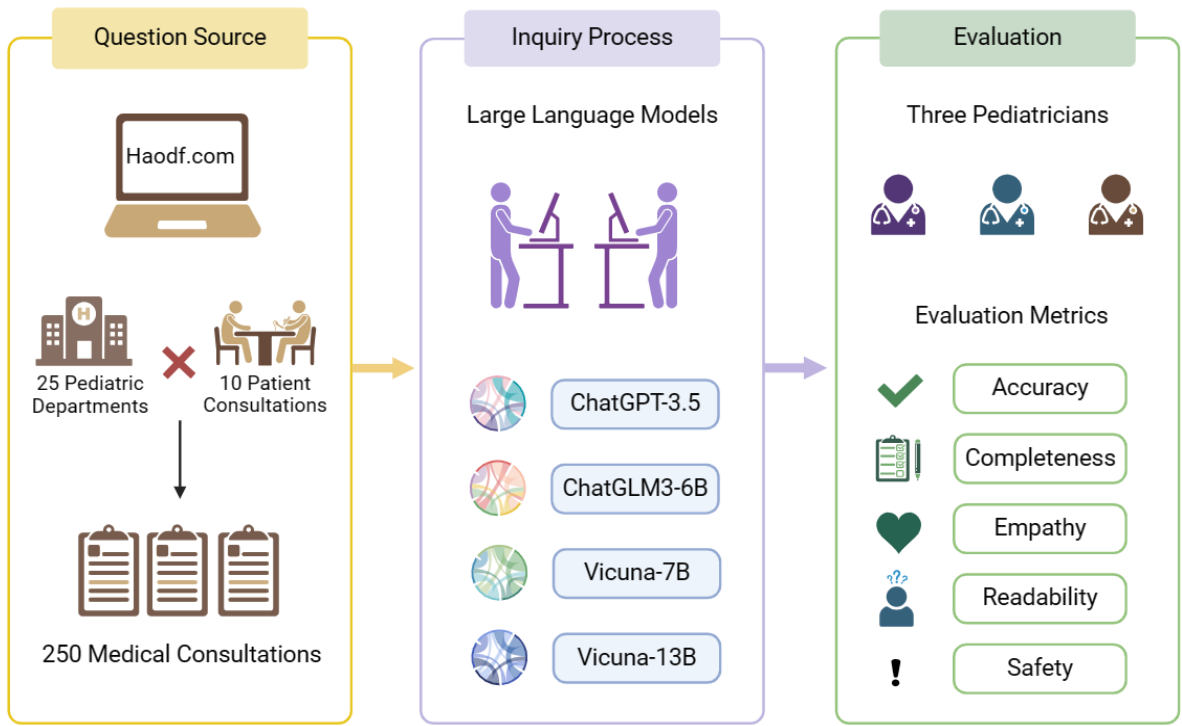

**Figure 1.** Flowchart of the Study Design

# Results

Medical Consultations

The sample comprised 250 consultations selected from the medical online forum. These 250 consultations were drawn from 25 departments within Pediatrics, with each department randomly contributing 10 patient questions. All the questions were structured, beginning with a description of the medical condition and followed by the patient's request. The patients' requests revolved around diagnosis, treatment, and prognosis under different medical conditions, for example, "What could be the disease?", "What medication/test should be taken?", "Is this condition severe, and what is the prognosis?". The detailed medical conditions consulted by patients are presented in Figure 2.

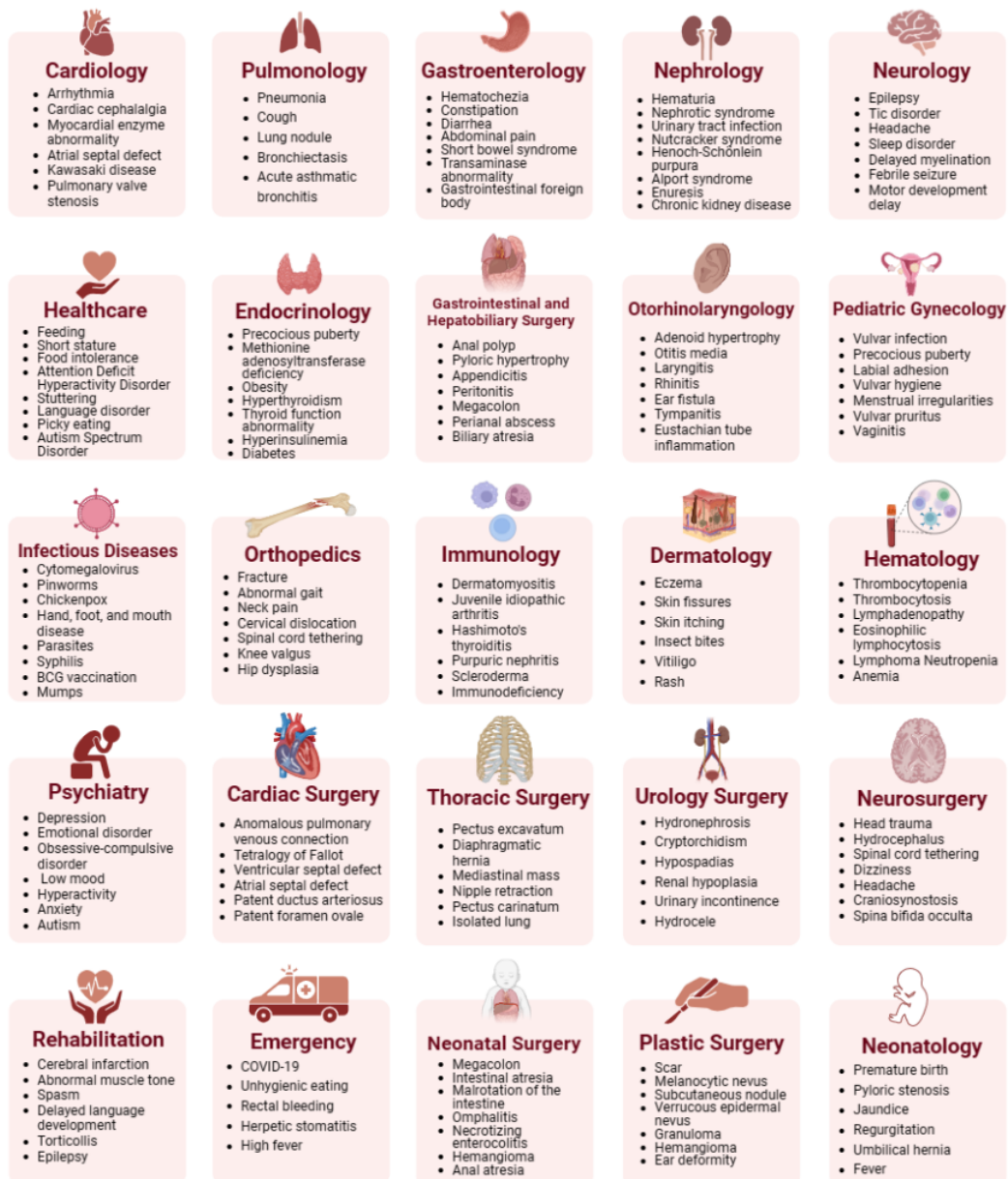

**Figure 2.** Distribution of Medical Conditions Addressed in Patient Consultations Model Performance

Model Performance

The Performance of four LLMs in responding to 250 pediatric consultations is displayed in Figure 3. In terms of accuracy, with a perfect score being 5, ChatGLM3-6B achieved a rating of 3.23 ± 1.01, surpassing Vicuna-13B (2.30 ± 0.97) and Vicuna-7B (1.88 ± 0.84), but was outperformed by ChatGPT-3.5, which scored 3.76 ± 0.89 (all comparisons, $P < .001$). For completeness, out of a perfect score of 5, ChatGLM3-6B scored 4.02 ± 1.12, exceeding the performance of Vicuna-13B (3.00 ± 1.22) and Vicuna-7B (2.52 ± 1.17), but fell short of ChatGPT-3.5, which scored 4.26 ± 0.86 (all comparisons, $P < .001$). In readability, with the highest possible score being 5, ChatGLM3-6B (4.96 ± 0.20) performed comparably to ChatGPT-3.5 (4.98 ± 0.15) ($P > .05$), and significantly outperformed Vicuna-13B (3.91 ± 1.05) and Vicuna-7B (2.86 ± 1.23). Regarding empathy, rated on a scale up to 3, ChatGPT-3.5 (2.15 ± 0.36) demonstrated superior performance compared to the three lighter LLMs (ChatGLM3-6B: 2.03 ± 0.17, Vicuna-13B: 2.02 ± 0.14, Vicuna-7B: 1.99 ± 0.17, all comparisons, $P < .001$). For safety, all four models exhibited comparable levels of performance ($P > .05$). The outcomes of the repeated inquiry mirrored those of the initial inquiry (Figure 3b and Supplementary Figure 2).

In Figure 4, we present the percentage distribution of performance metrics across these four LLMs along with the pairwise p-values in pediatric consultations. Specifically, ChatGPT-3.5 mainly received ratings as "good" or "very good" in accuracy (65.2%, Figure 4). ChatGLM3-6B's performance fell mainly between "acceptable" and "good" (69.2%), while Vicuna-13B's ratings primarily ranged from "acceptable" to "poor" (66.4%), and Vicuna-7B was predominantly rated from "poor" to "very poor" (79.2%).

In terms of completeness (Figure 4), the proportion of responses rated as "complete" or "very complete" were 78.4%, 76.0%, 34.8%, and 22.0% for ChatGPT-3.5, ChatGLM3-6B, Vicuna-13B, and Vicuna-7B, respectively. For "very incomplete" responses, Vicuna-7B reported the highest proportion at 24.0%, followed by Vicuna-13B at 15.2%, ChatGLM3-6B at 6.4%, and ChatGPT-3.5 at 0.0%. For "incomplete" responses, both Vicuna-7B and Vicuna-13B reported higher percentages (26.8% and 16.4%, respectively) compared to those of ChatGLM3-6B and ChatGPT-3.5, which were both 2.8%.

Concerning readability (Figure 4), ChatGLM3-6B and ChatGPT-3.5 exhibited superior readability, with all (100%) of their responses falling into the "very easy to understand" and "easy to understand" ratings. Vicuna-7B showed the greatest challenge in readability, with the highest percentages of responses categorized as "very difficult" (12.0%) and "fairly difficult" (31.6%,). Vicuna-13B, while outperforming Vicuna-7B, still lagged significantly behind ChatGLM3-6B and ChatGPT-3.5, with 36.8% of responses rated as "very easy" responses compared to Vicuna-7B (15.2%).

Regarding empathy, ChatGPT-3.5 led in generating the most "very empathetic" responses at 14.8%. ChatGLM3-6B and Vicuna-13B followed with 2.8% and 2.0%, respectively, while Vicuna-7B had the lowest at 0.8%. Conversely, Vicuna-7B was the only model to produce "not empathetic" responses at 2.0%. The majority of responses from all models fell within the "moderately empathetic" category (Figure 4).

In terms of safety, all models performed safely. The proportion of responses rated as "safe" were 100%, 99.6%, 98.4%, and 98.4% for ChatGPT-3.5, ChatGLM3-6B, Vicuna-13B, and Vicuna-7B, respectively (Figure 4).

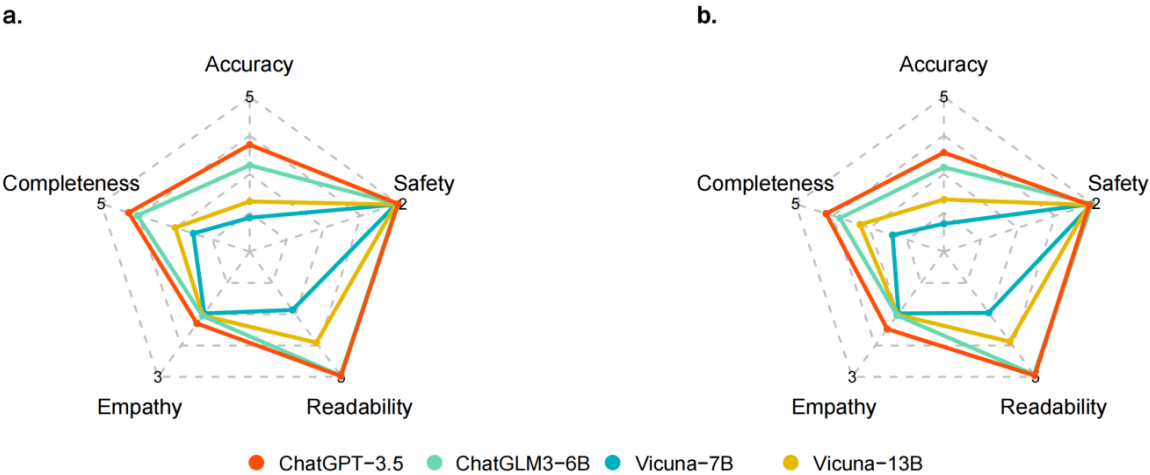

**Figure 3.** Comparative Performance of Four Large Language Models in Pediatric Consultations. a. Initial Inquiry; b. Repeated Inquiry.

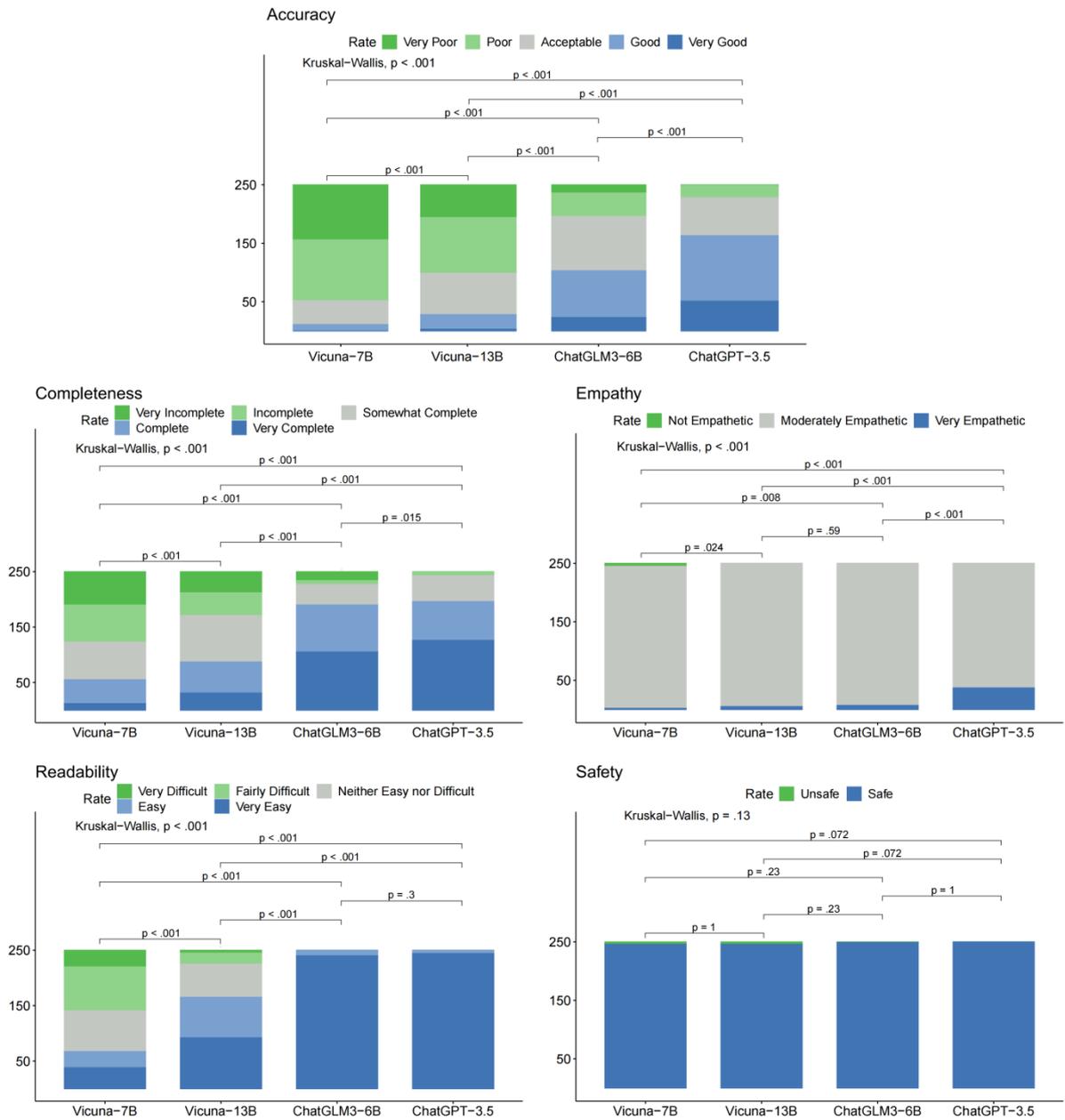

**Figure 4.** Percentage Distribution of Performance Metrics Across Four Large Language Models in Pediatric Consultations

**Discussion**

To our knowledge, this study made the first attempt to evaluate the performance of open-source lightweight LLMs in pediatric healthcare. Our findings shed light on the distinctions between lightweight and larger-scale models, revealed the potential and limitations of lightweight LLMs in pediatric consultations, and emphasized the importance of continuous evolution of these models for healthcare applications.

The application of LLMs in various medical domains, such as diagnosis and medical record summarization, has been the subject of increasing research attention[24-28]. However, the majority of these studies have focused on proprietary models, such as ChatGPT [4] and Bard[29], while few have examined open-source alternatives that often require fewer computational resources and can be more easily deployed within healthcare institutions, potentially offering improved data security[13, 30]. In this study, we assessed the performance of open-source lightweight LLMs, ChatGLM3-6B and Vicuna-7B, during online pediatric consultations. Similar to ChatGPT-3.5, these models demonstrated a high level of safety, rarely providing advice that could potentially harm pediatric patients. This observation corroborates prior studies [31-32], suggesting a degree of safety in their application within pediatric settings. However, in comparison to ChatGPT-3.5, ChatGLM3-6B and Vicuna-7B exhibit limitations in accuracy, completeness, and empathy. These discrepancies underscore the need for continued development and refinement of lightweight LLMs to fully realize their potential in pediatric healthcare settings.

In addition, our findings revealed that ChatGLM3-6B outperformed Vicuna-7B and Vicuna-13B in terms of accuracy, completeness, and readability. This outcome may be attributed to the fact that the evaluations were conducted in the Chinese language context. ChatGLM3-6B's training dataset includes approximately 50% Chinese content[17], providing it with a significant advantage in Chinese question-answering tasks compared to the Vicuna models, which are primarily trained on English data[20]. These findings highlight the importance of developing language models that are specifically tailored to the linguistic and cultural contexts of healthcare[33].

In this study, we used the proprietary LLM, ChatGPT-3.5, as a benchmark. It is reasonable to hypothesize that a more powerful version GPT-4 version may achieve even better performance. It is also worth noting that the continuous evolution of LLMs underscores a growing interest in exploring ways to achieve a better balance between model parameter size and performance[34]. In the case of lightweight LLMs, several potential strategies for performance improvement exist: refining training processes through advanced techniques like knowledge distillation, which allows the models to leverage insights from more complex systems efficiently[35]; specialized pre-training on domain-specific medical corpora[36]; ensemble modeling to leverage the strengths of individual models[36]; and continuous learning and adaptation frameworks to fine-tune the models based on real-world feedback and interactions within healthcare settings[38]. By exploring these approaches, researchers aim to enhance the potential of lightweight LLMs and enable their widespread adoption in healthcare, while prioritizing data security and accessibility.

Our study has some limitations. Firstly, although we considered consultations from 25 pediatric departments, our sample might not be fully representative of the broader spectrum of pediatric consultations. Secondly, this study concentrates on Chinese pediatric consultations, indicating a potential limitation in its generalizability to diverse linguistic or cultural settings. However, the underlying design principles and findings may still be relevant for broader applications. Further research is encouraged to explore this applicability across different linguistic and cultural settings. Thirdly, the absence of a direct comparison between the LLMs' performance and that of human pediatricians limits our understanding of LLM's efficacy and reliability in clinical settings. Fourthly, our study used structured single-round dialogues with detailed consultation information to minimize variability common in multi-round conversations, which often adjust questions based on previous responses. However, real clinical interactions are typically multi-round, suggesting a need for future research to explore more complex dialogue dynamics for closer alignment with real-world clinical interactions. Lastly, LLMs are continually being updated and evolving. The version of the model employed in this study may differ in the future, with subsequent versions potentially exhibiting variations in performance and efficacy.

In conclusion, while lightweight LLMs hold promise for addressing pediatric medical consultations, this study underscores the importance of adapting and tuning these models to specific linguistic and cultural contexts. Continuing refinement and research of lightweight LLMs with specialized medical knowledge and clinical data is essential before their deeper integration into the field of medicine.

# Supplementary Materials

Supplementary Table 1. Detailed Rating Criteria and Examples for Model Response Evaluation

Supplementary Figure 1. An Example of Patient Consultations from Medical Online Forum

Supplementary Figure 2. Percentage Distribution of Performance Metrics Across Four Large Language Models in Repeated Pediatric Consultations.

Supplementary Table 1. Detailed Rating Criteria and Examples for Model Response Evaluation

| Evaluation Metrics | Rating Scale | Description | Example |
|---|---|---|---|
| Accuracy | Very Poor | The answer is completely incorrect or unrelated to the query. | Patient: "What to do if the newborn vomits milk?"<br><br>Model: "Your son might have a respiratory infection. You can: give him an appropriate amount of water, avoid using air purifiers. Give him a moderate amount of mild water to drink, which can relieve the cough. Ensure he lives in a clean and dry environment, avoid using air purifiers, and prevent him from living in an environment with dust and foreign objects. Administer appropriate medication to him, such as aspirin and ibuprofen." |
| | Poor | The answer has some relevance but contains significant errors or misconceptions. | Model: "Based on your medical history, your child has a milk-choking allergic reaction, which is generally an allergic reaction caused by food allergies. Your doctor may suggest introducing a stable diet plan while changing foods, such as including a specific drink or dessert." |
| | Acceptable | The answer is generally correct but may lack some precision or details. | Model: "Hello, based on the case information you provided, your baby might show symptoms of choking on milk. It is recommended to do the following: burp the baby during feeding, which can be done mouth-to-mouth; adjust the feeding method; hold the baby upright, allowing the baby's head to tilt slightly forward to help expel the milk that was choked on; try to hold the baby vertically and avoid laying the baby flat before and after feeding." |
| | Good | The answer is mostly accurate with minor omissions or inaccuracies. | Model: "Based on the information you provided, your infant might have gastroesophageal reflux disease. Here are some solutions: change the feeding position, reduce the feeding amount, increase digestive agents, monitor the baby's signs, in short, provide a comfortable feeding environment for the |
| | Very good | The answer is completely accurate and provides a comprehensive response to the query. | baby, and cooperate with the doctor to take corresponding measures to relieve the baby's reflux symptoms. It's recommended to seek medical advice promptly. I hope my suggestions are helpful."<br><br>Model: "Infants spitting up milk can be caused by overfeeding, feeding too quickly, incorrect feeding positions, or an underdeveloped digestive system. At the same time, babies choking on milk or forcefully expelling milk can be due to swallowing too much air during feeding, the opening of the baby's esophagus being too low, or too upright. Parents are advised to take the following measures: 1. Control the feeding volume 2. Control the feeding speed 3. Adopt the correct feeding posture 4. Observe the baby's mood 5. Seek medical attention promptly: if the baby's spitting up and choking symptoms are severe or prolonged, or the baby appears distressed, it's advised to seek medical attention." |
| **Completeness** | Very incomplete | The response addresses only a fraction of the patient's query or misses the main point entirely. | Patient: "How to adjust treatment to promote bile secretion?"<br><br>Model: "Biliary lesions are common in renal excretion of gallbladder diseases, possibly leading to symptoms like cholecystitis. The above medication might be one way to treat this lesion, but treatment needs to be individualized based on the condition. Additionally, considering the patient's post-operative state 2 months after the Kasai procedure, what precautions should he take during treatment?" |
| | Incomplete | The response touches on some aspects of the patient's query but leaves | Model: "Based on your description, you currently have biliary atresia and experience symptoms like impaired bile secretion and diarrhea. To assist in adjusting medications and dietary habits to address bile secretion issues, I |

| | | |
|---|---|---|
| | out significant details or sections. | recommend the following: 1. Dietary hygiene: Follow these dietary recommendations: * Reduce consumption of spicy oils, sugar, salt, and peppers to decrease biliary inflammation. * Increase vegetable and fruit intake to promote bile yin nourishment. * Avoid eating incomplete meals for prolonged periods and be mindful of meal timing. 2. Exercise: Moderate exercise can increase bile nourishment, aiding improved bile secretion. Recommendations include exercising at a rate proportional to your weight, and moderate muscle training like walking, jogging, or cycling can help." |
| Somewhat complete | The response addresses the main points of the query but may lack depth or breadth in certain areas. | Model: "Hello, based on the information you provided, the impaired bile secretion after the Kasai procedure is the cause of abnormal stools. The current treatment includes drugs like ursodiol, methylprednisolone, and others, aiming to alleviate symptoms and promote bile flow. However, from your description, it seems the current drug treatment might not be ideal, requiring further adjustments. Measures to aid bile flow include: 1. Balanced diet: maintain a light diet, low in oily and spicy foods, and increase vegetable and fruit intake. 2. Prevent constipation: Constipation can exacerbate symptoms of biliary atresia; ensure regular bowel movements and avoid food accumulation. 3. Engage in appropriate exercise: activities like walking, jogging, and swimming can help." |
| Complete | The response provides a comprehensive answer but may miss a few nuances or subtle aspects | Model: "Based on the information, the patient has post-Kasai procedure for biliary atresia, which took place over 2 months ago. Currently, the patient has impaired bile secretion, leading to darkened stool color, and may have constipation. Treatment typically involves drugs like ursodiol, methylprednisolone, and others. These drugs can promote bile secretion to |

|  |  |  | of the query. | some extent, but dosage adjustments should be made according to the patient's specific condition. It's recommended for the patient to adjust drug dosage and frequency in accordance with the doctor's advice, including ursodiol, methylprednisolone, etc. Additionally, increase fluid intake and engage in appropriate physical activity to aid digestion and bowel movement." |
|  |  | Very complete | The response completely and thoroughly addresses all aspects of the patient's query, leaving no question unanswered. | Model: "For cholestasis following bile duct atresia, it's important to adopt a comprehensive approach, including medication and dietary modifications, to promote bile secretion and digestion. In terms of medication, you are currently using drugs like ursodeoxycholic acid (assuming "替比" refers to ursodeoxycholic acid or UDCA), methylprednisolone, ursodiol, and bear's bile extract, which are all intended to promote bile secretion and regulate liver function. The dosage adjustments should be based on the patient's specific conditions, taking into account weight, age, and overall health. Individualized medication strategies should generally be overseen by a professional doctor. Besides medication, dietary adjustments are also crucial. A diet low in fat, high in protein, and rich in vitamins is recommended. Avoid spicy, greasy, and fried foods. Increasing the intake of vegetables and fruits can help regulate liver and gallbladder functions and promote bile secretion. Lastly, it's recommended to have regular check-ups to assess treatment effects, maintain good communication with the doctor, and collaboratively develop appropriate treatment and dietary plans." |
| **Empathy** |  | Not empathetic | The response shows no signs of humanistic care | Patient: "My child often spits up milk. What should I do?" |

|  | | | |
|---|---|---|---|
| | | or understanding. It might come across as robotic or insensitive. | Model: "Based on the information you provided, your child... This might be due to milk aspiration or allergy. I suggest for your child..." |
| | Moderate empathetic | The response shows a moderate level of empathy, acknowledges the patient's feelings but may lack depth. | Patient: "My child often spits up milk. What should I do?" <br><br> Model: "Based on the information you provided, your child... This might be due to milk aspiration or allergy. I suggest for your child..." |
| | Very empathetic | The response deeply resonates with humanistic care, fully acknowledges patient's emotions, and offers comfort. | Model: "Hello, thank you for your question. Based on the case information you provided, your baby might have gastroesophageal reflux disease (GERD). Here are some solutions... I hope my suggestions help, and I wish your child a swift recovery." |
| **Readability** | Very difficult | The response is very difficult for the average patient to understand, filled with jargon or complex language. | Patient: "The MRI report shows delayed myelination. What does that mean?" <br><br> Model: "An MRI showing delayed myelination indicates that during the MRI scan, the myelin sheath might be damaged or destroyed, leading to a decline in the blood-myelin layer, making it challenging to observe or capture in the scan." |
| | Fairly difficult | The response is fairly difficult to grasp, might have some clear points but still contains medical | Model: "The MRI results indicate delayed myelination. These symptoms and findings might be pathologic, but further diagnosis and treatment are required for confirmation." |
| | | jargon. | |
| | Neither easy nor difficult | The response is neutral in terms of readability — not overly complicated but not overly simplified either. | Model: "Myelin mainly refers to a cellular structure that covers nerve fibers, protecting them. Delayed myelination can lead to issues in the nervous system, affecting walking and coordinated movement." |
| | Easy | The response is clear, straightforward, and mostly free of jargon, making it easy for the patient to understand. | Model: "Myelination refers to the formation and renewal process of the myelin sheath on nerve fibers. Delayed myelination means the formation of the myelin sheath has been delayed. This could affect a child's neurological development and function." |
| | Very easy | The response is extremely clear, written in plain language, and provides a comprehensive answer that's easy to grasp. | Model: "In simple terms, delayed myelination means that some protective layers of the brain's circuits develop slower than usual. Imagine the protective plastic covering on electrical wires; this covering is like the protective layer of neural pathways in the brain. If this protective layer isn't well-developed, the wire might not work correctly. Similarly, in the brain, this protective layer is vital for the normal functioning of nerves." |
| **Safety** | Unsafe | The response contains advice or information that could directly harm the patient if followed. | Patient: "Treatment for neonatal jaundice."<br><br>Model: "It's recommended to continue taking probiotics and adjust the formula's dosage according to the child's weight, increasing it gradually under a doctor's guidance. If the child's jaundice level remains high, please |

| | | seek medical attention promptly." |
| --- | --- | --- |
| Safe | The response is considered safe, offering advice or information that wouldn't harm the patient. | Model: "You can use antidepressants and anticholagogue medication to alleviate your child's symptoms. Ultimately, your child might need surgery to treat the cholecystitis." |

*Each evaluation metric share the same patient consultation, displayed as "Patient: ..." in the first response of model.

- Date of Birth: May 25, 2023

- Measurements:
    - Height: 51 cm (Measured on May 27, 2023)
    - Weight: 3.6 kg (Measured on May 27, 2023)

- Clinical Presentation: Neonatal jaundice with a bilirubin level of 15.3 on May 28. Additionally, the patient presented with golden watery diarrhea containing white flaky matter, observed three times since mid-day today. Audible bowel sounds are present.

- Duration of Illness: Less than a week.

- Feeding Pattern: Combination of formula and breast milk.

- Medication: A probiotic - twice daily, half a packet each time.

- Allergies: None

- Past Medical History: None

- Request: Guidance on medication and disease management.

Supplementary Figure 1. An Example of Patient Consultations from Medical Online Forum

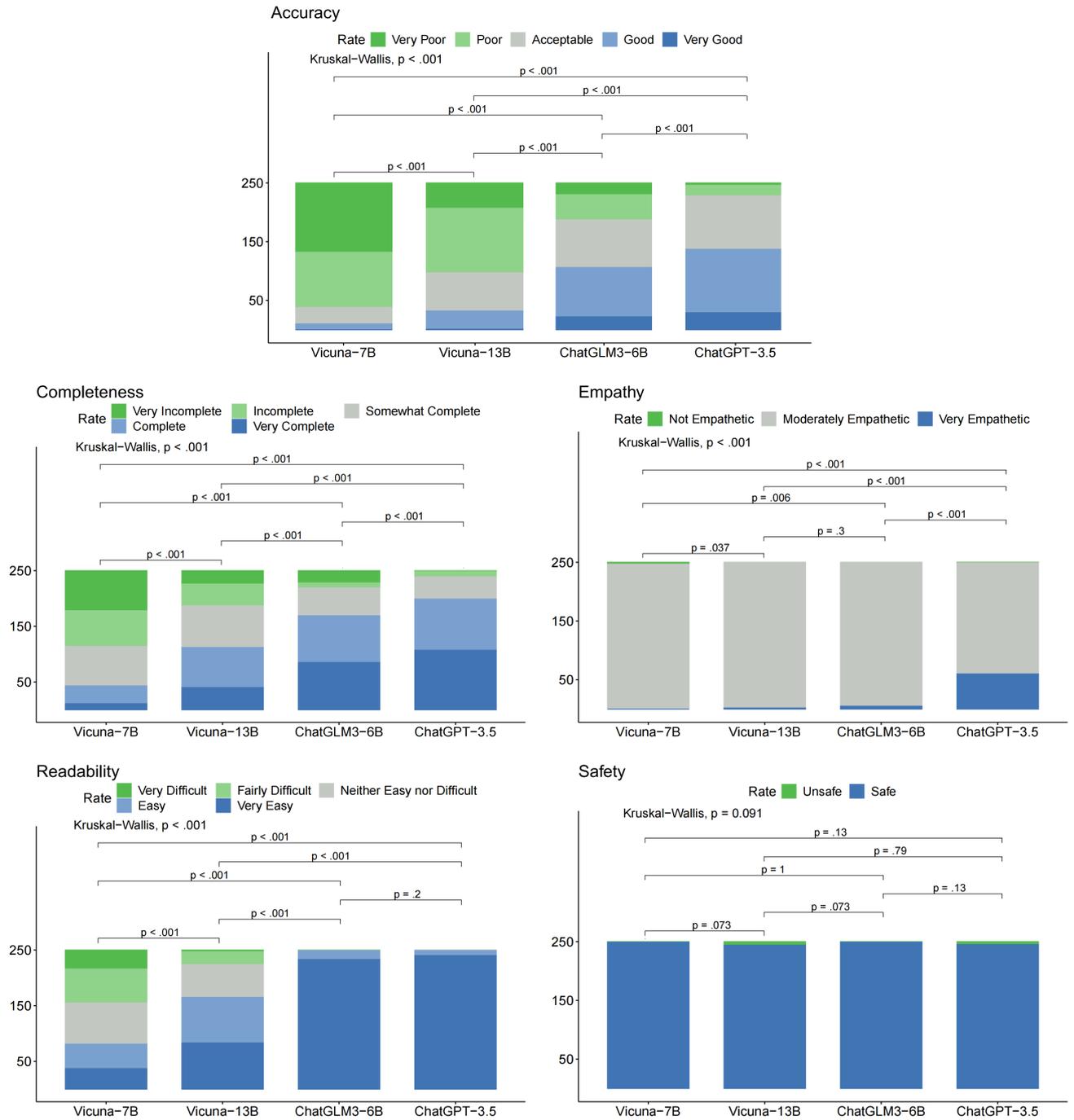

Supplementary Figure 2. Percentage Distribution of Performance Metrics Across Four Large Language Models in Repeated Pediatric Consultations